\documentclass[10pt]{article} 

\usepackage[preprint]{tmlr}

\usepackage{graphicx} 
\usepackage{hyperref}
\usepackage{url}
\usepackage{subcaption}
\usepackage{cleveref}
\usepackage{epigraph}  
\usepackage{xcolor}
\usepackage{tikz}
\usetikzlibrary{3d}
\usepackage{caption}
\usepackage{subcaption}
\usepackage{booktabs}
\usepackage[numbers,compress,sort]{natbib}

\title{Improving Generalization on the ProcGen Benchmark\\ with Simple Architectural Changes and Scale}
\author{\name Andrew Jesson 
    \email adj2147@columbia.edu \\
    \addr Department of Statistics\\ Columbia University
    \AND
    \name Yiding Jiang 
    \email yidingji@cs.cmu.edu \\
    \addr Machine Learning Department \\ Carnegie Mellon University
}

\ifdefined\nohyperref\else\ifdefined\hypersetup
    \definecolor{mydarkblue}{rgb}{0,0.08,0.45}
    \hypersetup{ %
        pdftitle={},
        pdfsubject={},
        pdfkeywords={},
        pdfborder=0 0 0,
        pdfpagemode=UseNone,
        colorlinks=true,
        linkcolor=mydarkblue,
        citecolor=mydarkblue,
        filecolor=mydarkblue,
        urlcolor=mydarkblue,
    }
    \fi
\fi

\begin{document}

\maketitle

\begin{abstract}
We demonstrate that recent advances in reinforcement learning (RL) combined with simple architectural changes significantly improves generalization on the ProcGen benchmark. These changes are frame stacking, replacing 2D convolutional layers with 3D convolutional layers, and scaling up the number of convolutional kernels per layer. Experimental results using a single set of hyperparameters across all environments show a 37.9\% reduction in the optimality gap compared to the baseline (from 0.58 to 0.36). This performance matches or exceeds current state-of-the-art methods. The proposed changes are largely orthogonal and therefore complementary to the existing approaches for improving generalization in RL, and our results suggest that further exploration in this direction could yield substantial improvements in addressing generalization challenges in deep reinforcement learning.
\end{abstract}


\epigraph{``\emph{The only thing that is constant is change.}''}{-- Heraclitus}

\section{Introduction}

Reinforcement learning (RL) research often focuses on improving performance in a fixed environment, but real world models are rarely deployed in the exact environment they are trained in. 
For example, an RL algorithm for autonomous driving may only be trained on experiences collected from a limited number states, cities, or neighborhoods, but is required to generalize to similar but unseen environments and conditions.
Such generalization poses a significant challenge for current deep RL algorithms.
The ProcGen benchmark~\citep{cobbe2020leveraging} is the canonical benchmark for studying generalization in RL. 
ProcGen consists of 16 distinct \emph{environments} that visually resemble those in Atari~\citep{atari} and each environment has different procedurally generated \emph{levels}.
The levels are individually distinct from each other but share the same underlying strategies, and agents are constantly exposed to new, unseen levels during both training and evaluation. 
This difficult setting tests the model's ability to acquire generalizing behaviors, rather than simply memorize specific level configurations.

In the past, several methods treated generalization in ProcGen as a representation learning problem and adapted various techniques from supervised deep learning \citep{cobbe2020leveraging, igl2019generalization, ye2020rotation,raileanu2020automatic}. 
More recently, \citet{jiang2024importance} demonstrated that a key bottleneck for generalization in RL is exploration, and showed that one can make a competitive value-based RL method through uncertainty-driven exploration. 
\citet{jesson2024relu} also found that similar principles can be applied to policy-based RL methods.
Their method, VSOP, incorporates advantage clipping and dropout into the reinforce algorithm to yield approximate Bayesian inference over the policy network's parameters, which enables state-aware exploration via Thompson sampling.
The algorithmic solutions in both approaches are quite general and are not restricted to ProcGen, which suggests that efficient exploration could be a key aspect of building generalizable reinforcement learning algorithms.

Most existing approaches share the same ResNet backbones for processing the video state-space \citep{schulman2017proximal,cobbe2021phasic,raileanu2021decoupling,jiang2024importance,jesson2024relu} and architectural changes are relatively under explored.
We argue that using a standard feed-forward architecture ignores some crucial aspects of the problem structure in RL generalization.
A key challenge in such environments is partial observability.
In many ProcGen environments, the agent's current observation may not fully encapsulate all the necessary information required to make optimal decisions, since different levels of an environment could have some shared one-step observations but vastly different dynamics, resulting in distinct optimal policies~\citep{raileanu2021decoupling}.
This means that the agent must integrate information over time, as relying solely on individual frames can result in suboptimal decisions.
Previous experiments in the ProcGen report have shown that incorporating a recurrent component like an LSTM can lead to improvement in a limited number of environments~\citep{cobbe2020leveraging}, but to the best of our knowledge, this direction has not been extensively explored.
This observation indicates the potential for methods that can effectively handle temporal dependencies in non-Markovian environments.

Toward this goal, we investigate using frame stacking and 3D convolutions to capture more temporal information directly from the observation space.
By stacking multiple consecutive frames, the model is given a richer temporal context, which helps mitigate the potential partial observability of the environment.
Moreover, 3D convolutions allow the model to extract spatial-temporal features from these stacked frames, which may also assist in more robust generalization across procedurally generated levels.
Lastly, we show that simply scaling the proposed architecture can significantly improve the performance further, similar to the observed benefits of computation scale in other fields of deep learning~\citep{hoffmann2022training}.

Our contributions include an empirical analysis of these techniques, as well as an exploration of scaling the model capacity to capture richer representation.
Initial results show promising improvements in generalization, with a 37.9\% reduction in the optimality gap (from 0.58 to 0.36) compared to the baseline VSOP algorithm \citep{jesson2024relu}.
This performance, to the best of our knowledge, is competitive with or even exceeds the current state-of-the-art methods for ProcGen.

\begin{figure*}[ht]
     \centering
     \begin{subfigure}[t]{0.43\textwidth}
         \centering
         \includegraphics[width=\textwidth]{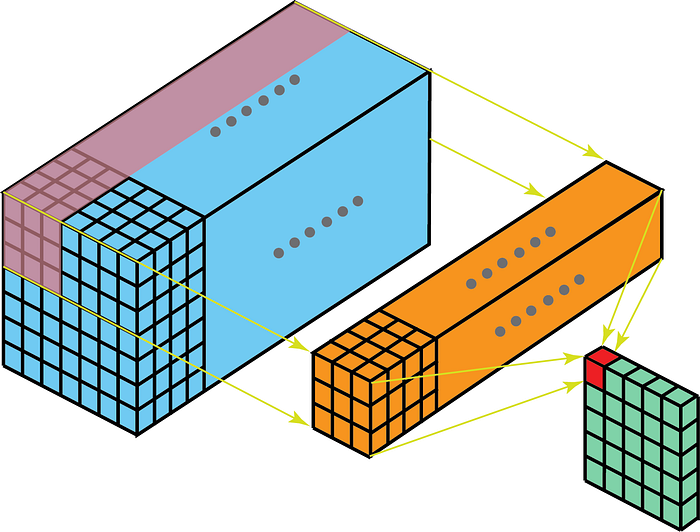}
         \caption{2D convolution}
     \end{subfigure}
     \hspace{12mm}
     \begin{subfigure}[t]{0.43\textwidth}
         \centering
         \includegraphics[width=\textwidth]{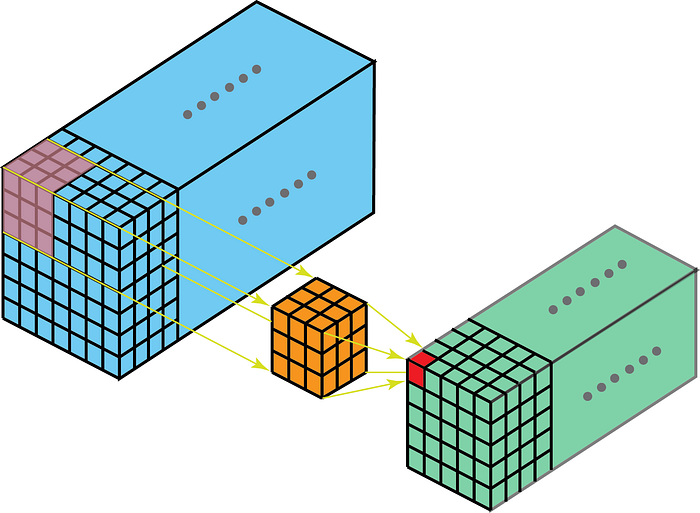}
         \caption{3D convolution.}
     \end{subfigure}
        \caption{
            Illustration of 3D convolution vs 2D convolution. 
            The 3D convolution is able to process the context at a finer temporal granularity whereas the 2D convolution must process all the stacked temporal information at the same time.
            The blue tensor is the input, the orange tensor is the convolutional kernel, and the green tensor is the resulting output.
            3D convolution can have multiple output channels too but it is not shown in this figure as visualizing a 4D tensor is very hard. 
            This visualization is created by user \texttt{ashenoy} at Stack Exchange (\url{https://ai.stackexchange.com/questions/13692/when-should-i-use-3d-convolutions}) under a CC BY-SA license.
        }
        \label{fig:3dconv}
\end{figure*}

\section{Methodology}
This study involves modifying VSOP---an on-policy deep reinforcement learning algorithm \citep{jesson2024relu}---by incorporating frame stacking, replacing 2D convolutions with 3D convolutions, and increasing the number of convolutional kernels.
These changes are straightforward and only involve changing the model architectures without any algorithmic changes.
We examine two different configurations.

\paragraph{VSOP-3D.} 
This variant is used to assess the effect of modeling spatiotemporal features by using 3D convolutions and stacking the last 8 frames to represent states while keeping all other hyperparameters set to the values specified for the unmodified VSOP algorithm.
Intuitively, 3D convolution is a better candidate for processing input with stacked frames because it can process different time segments independently (Figure~\ref{fig:3dconv}).
In comparison, the 2D convolution processes all stacked frames as densly connected channels of a single image, which omits inductive biases associated with temporal regularity.

\paragraph{VSOP-3D+.}
This variant is used to assess the effect of scale by increasing the number of frame stacking to 16 frames and doubling the number of convolutional channels in each layer.
Due to limited GPU memory, the number of mini-batches per epoch was increased from 8 to 32.
Due to limited GPU memory, we reduce the minibatch size from 2048 to 512.
To keep run time below cluster limits and the total number of model updates roughly constant, the number of updates per epoch was reduced from 3 to 1.
To counteract the effect of smaller mini-batches, the learning rate was reduced from $4.5 \times 10^{-4}$ to $2.0 \times 10^{-4}$. 

\begin{table}[ht]
    \centering
    \begin{tabular}{@{}lccccc@{}}
        \toprule
        \textbf{Hyperparameter} & \textbf{PPO} & \textbf{PPO-3D} & \textbf{VSOP} & \textbf{VSOP-3D} & \textbf{VSOP-3D+} \\
        \midrule
        Number of frames & 1 & 8 & 1 & 8 & 16 \\
        Width multiplier & 1 & 1 & 1 & 1 & 2 \\
        Learning rate &  $5 \times 10^{-4}$ &  $5 \times 10^{-4}$ &  $4.5 \times 10^{-4}$ &  $4.5 \times 10^{-4}$ & $2.0 \times 10^{-4}$ \\
        Batch size & 2048 & 2048 & 2048 & 2048 & 512 \\
        Epochs per update & 3 & 3 & 3 & 3 & 1 \\
        $\gamma$ & 0.999 & 0.999 & 0.999 & 0.999 & 0.999 \\
        GAE-$\lambda$ & 0.95 & 0.95 & 0.881 & 0.881 & 0.881 \\
        Norm. advantages & True & True & N/A & N/A & N/A \\
        Clip value loss & True & True & N/A & N/A & N/A \\
        Clip coeff. & 0.2 & 0.2 & N/A & N/A & N/A \\
        Entropy coefficient & $1 \times 10^{-2}$ & $1 \times 10^{-2}$ & $1 \times 10^{-5}$ & $1 \times 10^{-5}$ & $1 \times 10^{-5}$ \\
        Value loss coeff. & 0.5 & 0.5 & 0.5 & 0.5 & 0.5 \\
        Max gradient norm & 0.5 & 0.5 & 0.5 & 0.5 & 0.5 \\
        Dropout rate & N/A & N/A & 0.075 & 0.075 & 0.075 \\
        \bottomrule
    \end{tabular}
    \caption{
        Hyperparameters for PPO, PPO-3D, VSOP, VSOP-3D, and VSOP-3D+. 
        PPO and PP0-3D user the baseline PPO hyperparameters \cite{cobbe2020leveraging,huang2022cleanrl}. VSOP and VSOP-3D use the baseline VSOP hyperparameters \cite{jesson2024relu}. 
        VSOP-3D+ uses a reduced batch-size, number of epochs per update, and learning rate to account for resource constraints.
    }
\end{table}

\section{Results}

\begin{figure}[ht]
    \centering
    \includegraphics[width=0.98\linewidth]{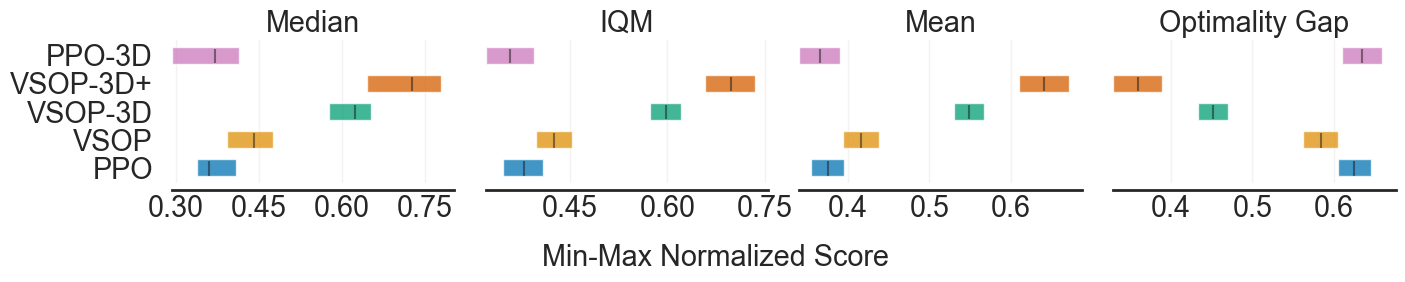}
    \caption{
        \textbf{Efficacy.}
        Aggregate test level evaluation metrics for average episodic return over the last 100 steps.
        Compared to VSOP, VSOP-3D is given 8 frames instead of 1 frame and uses 3D instead of 2D convolutions.
        VSOP-3D+ scales VSOP-3D by increasing the number of frames to 16 and doubling the number of convolutional channels in each layer.
        We observe considerable improvements in all evaluation metrics considered. Comparing VSPO-3D+ to baseline VSOP, we observe a 65.9\% increase in the Median (from 0.44 to 0.75), a 62.8\% increase in the IQM (from 0.43 to 0.70), a 52.5\% increase in the Mean (from 0.42 to 0.64), and a 37.9\% decrease in the Optimality Gap (from 0.58 to 0.36).
    }
    \label{fig:enter-label}
\end{figure}

\begin{figure}[ht]
    \centering
    \begin{tabular}{cccc}
        \includegraphics[width=0.22\linewidth]{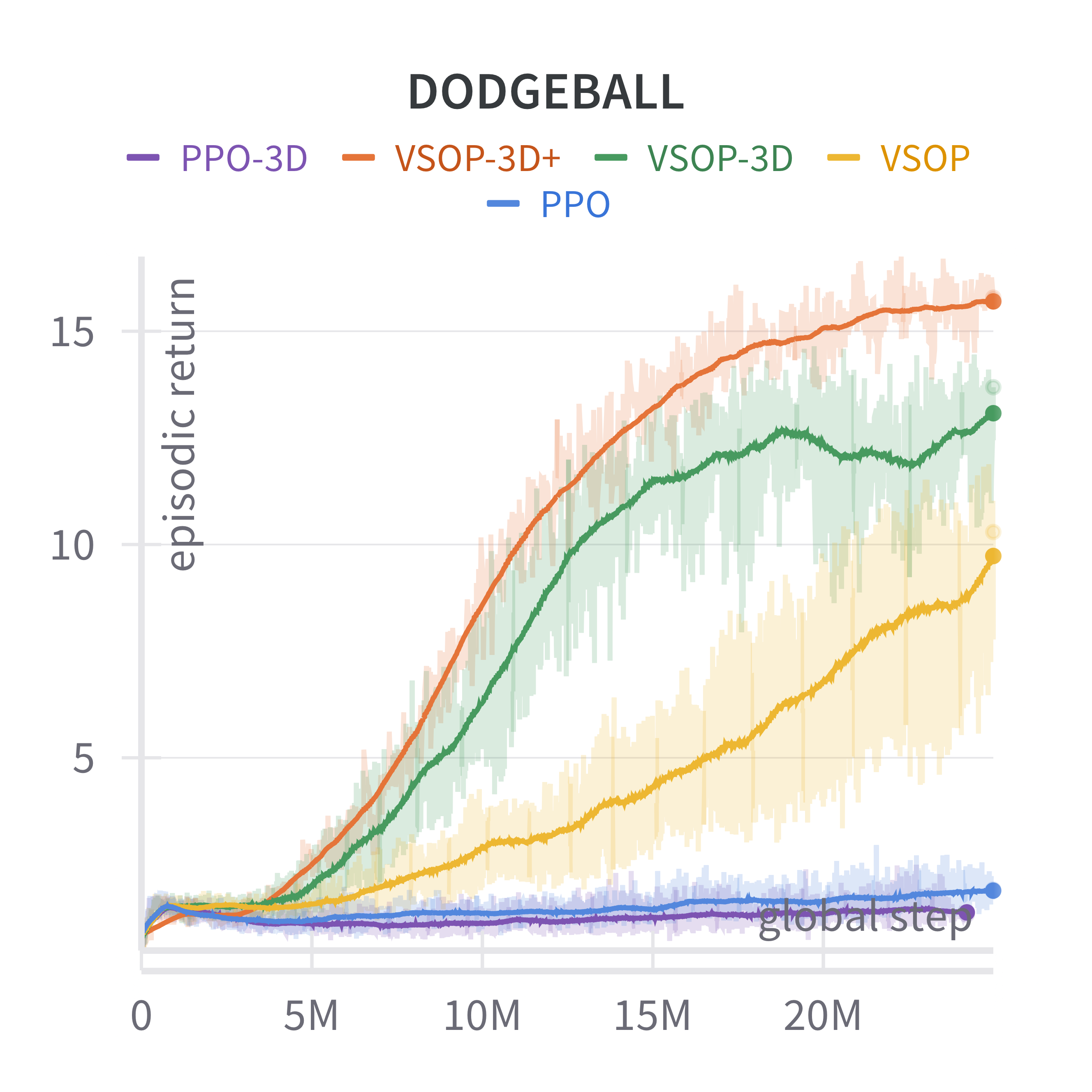} &
        \includegraphics[width=0.22\linewidth]{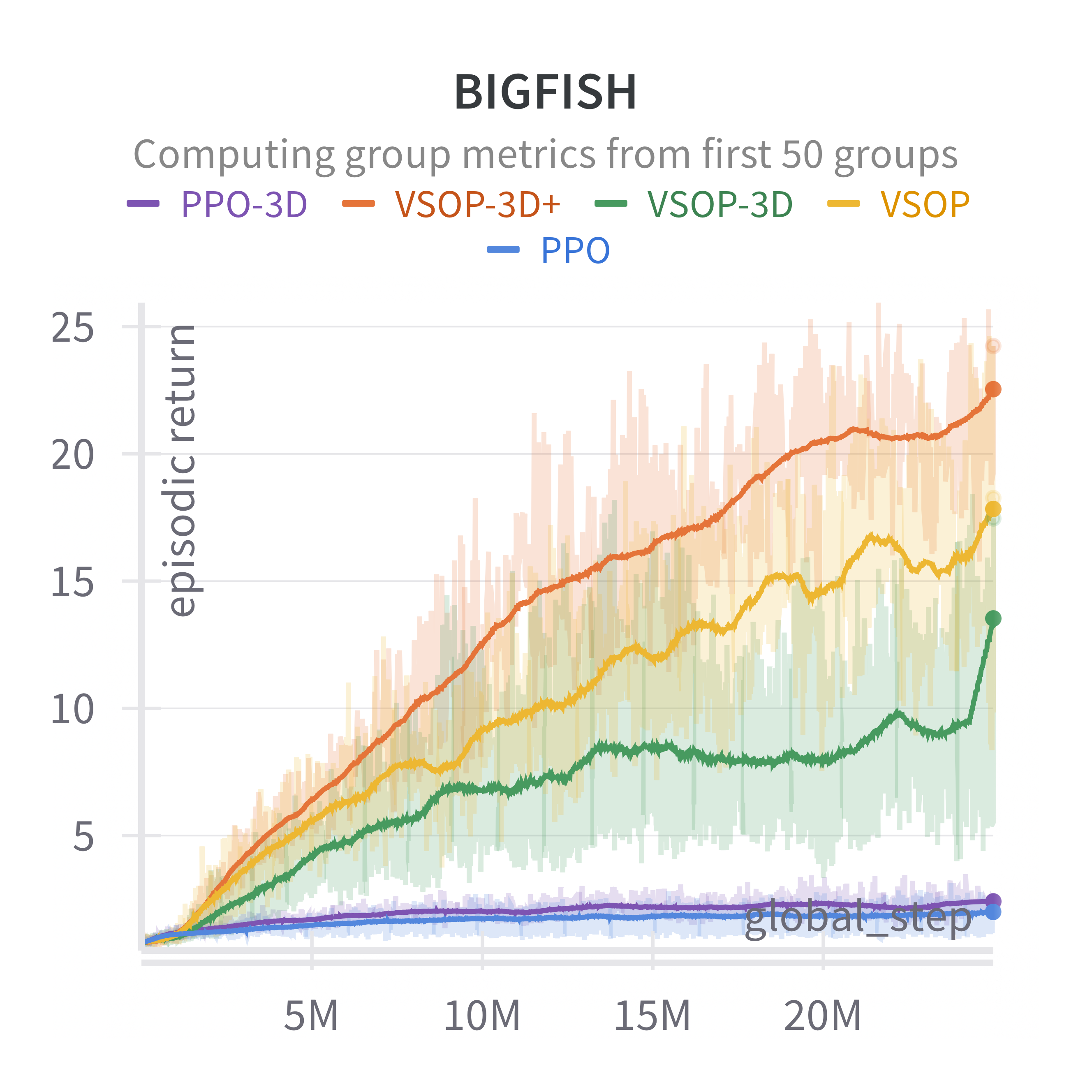} &
        \includegraphics[width=0.22\linewidth]{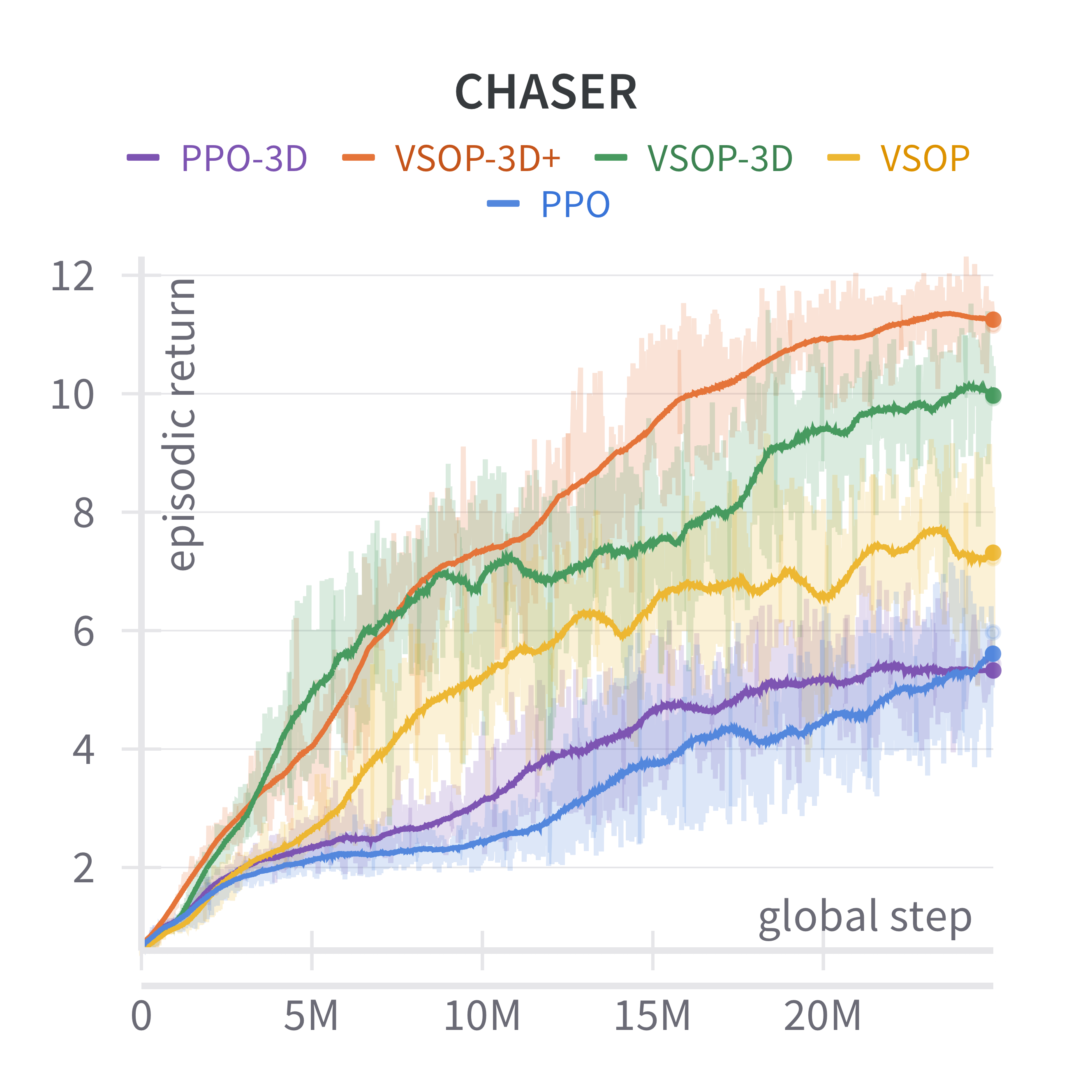} &
        \includegraphics[width=0.22\linewidth]{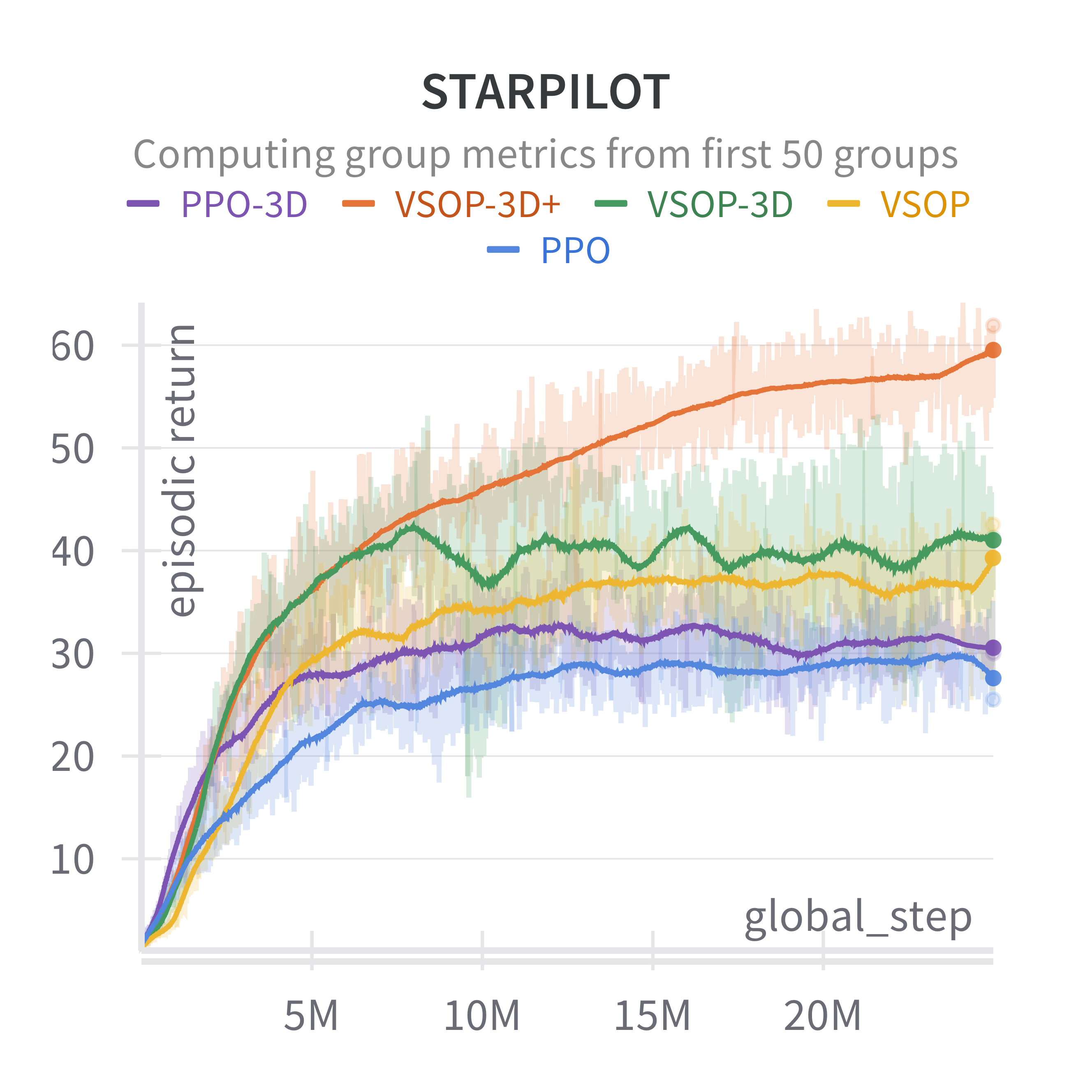} \\
        \includegraphics[width=0.22\linewidth]{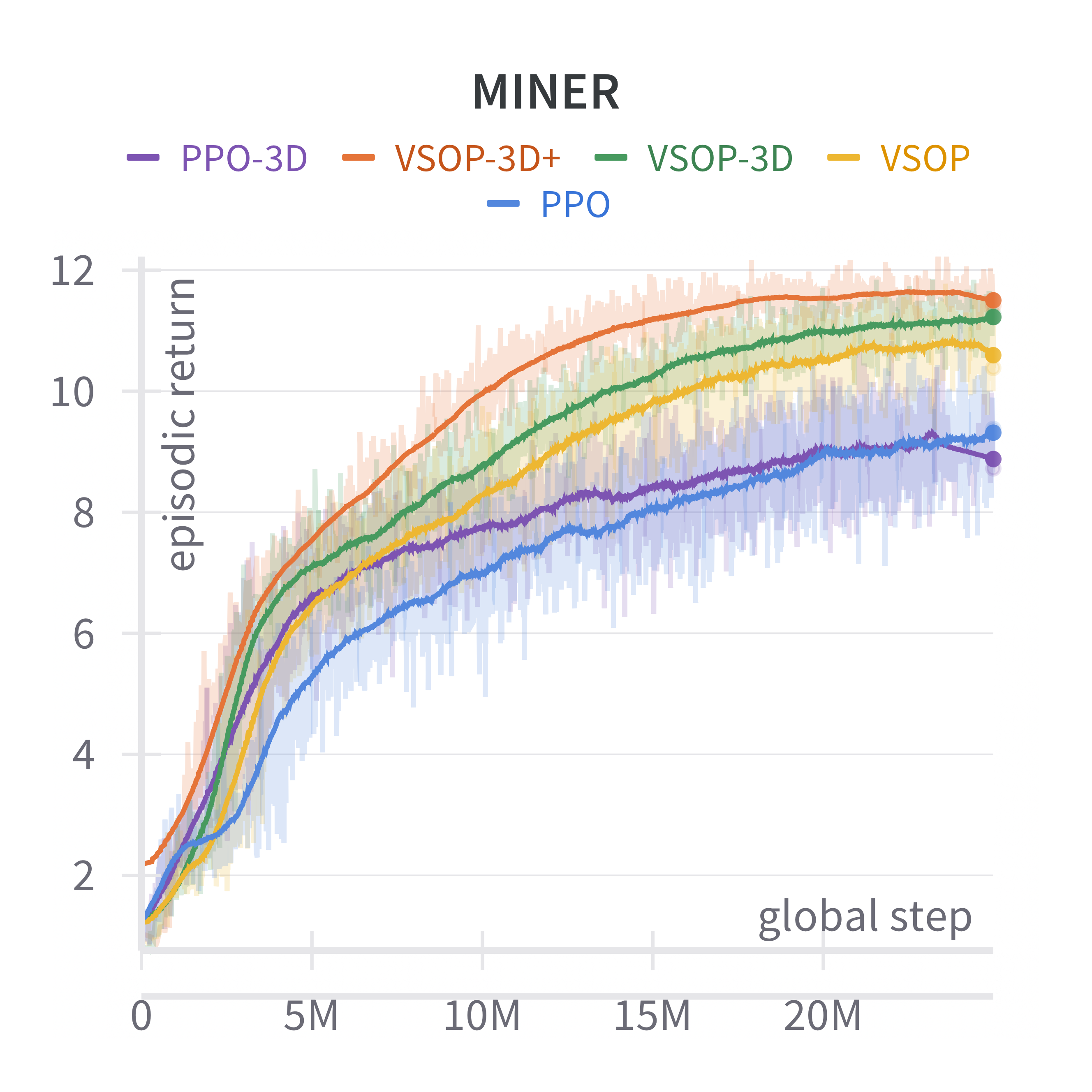} &
        \includegraphics[width=0.22\linewidth]{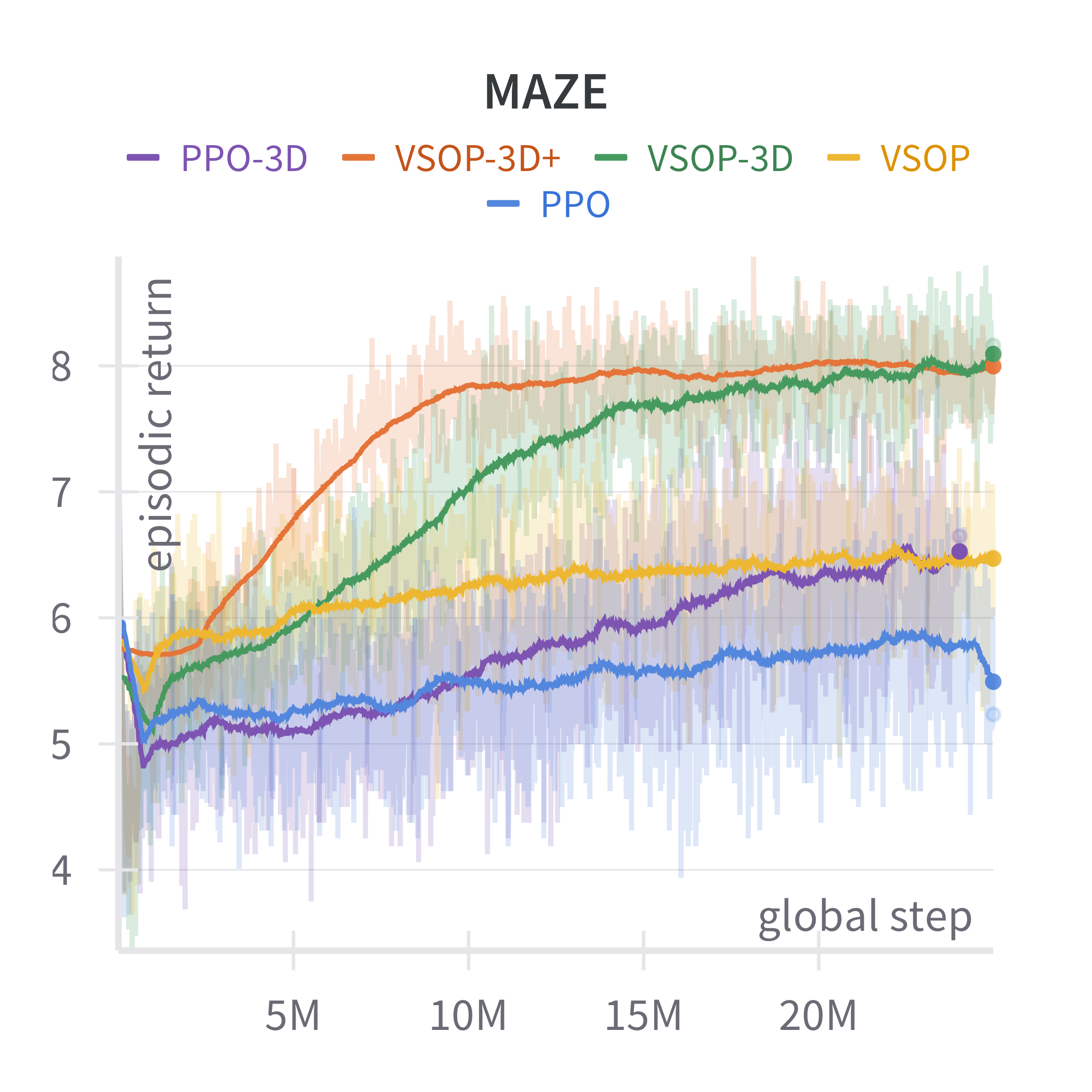} &
        \includegraphics[width=0.22\linewidth]{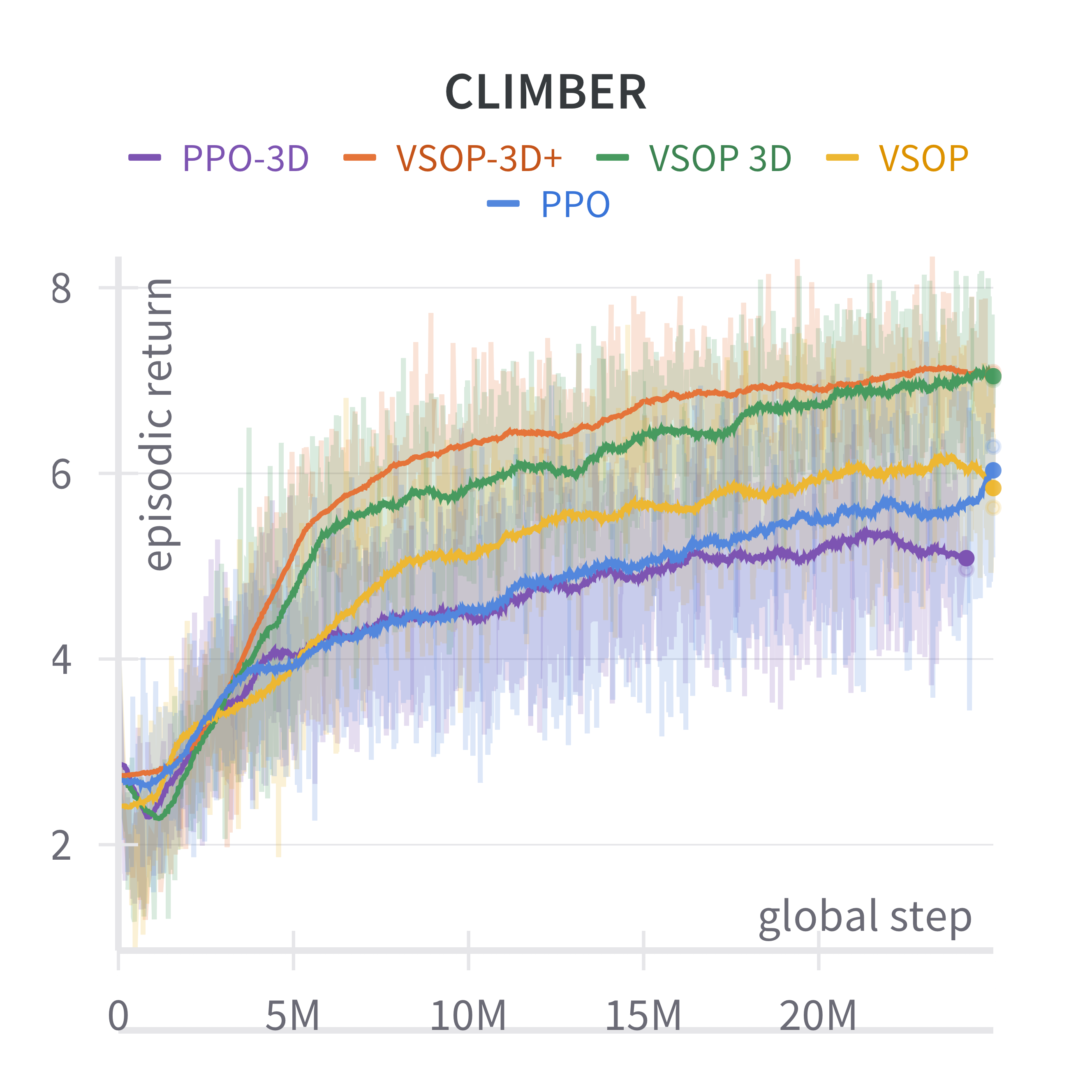} &
        \includegraphics[width=0.22\linewidth]{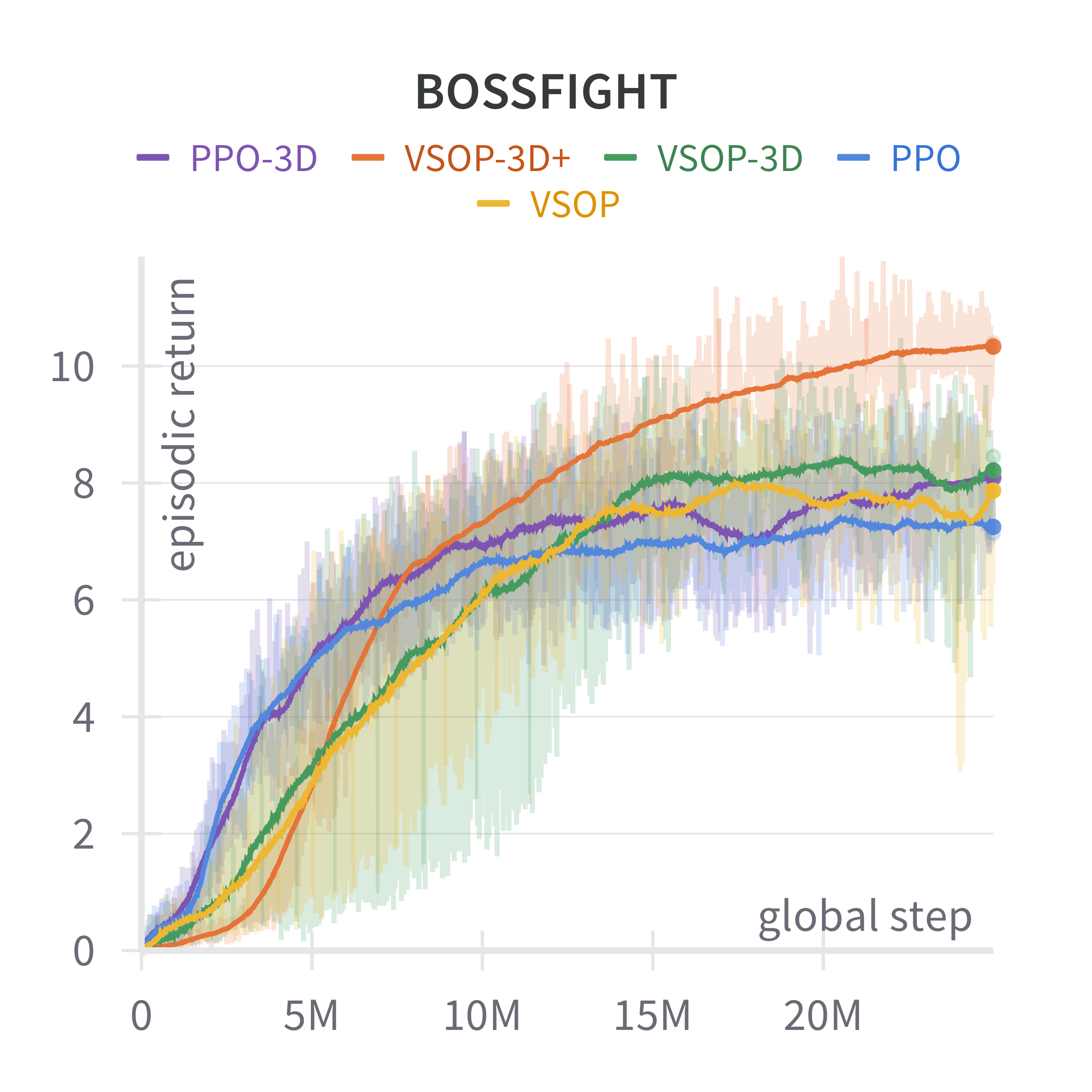} \\
        \includegraphics[width=0.22\linewidth]{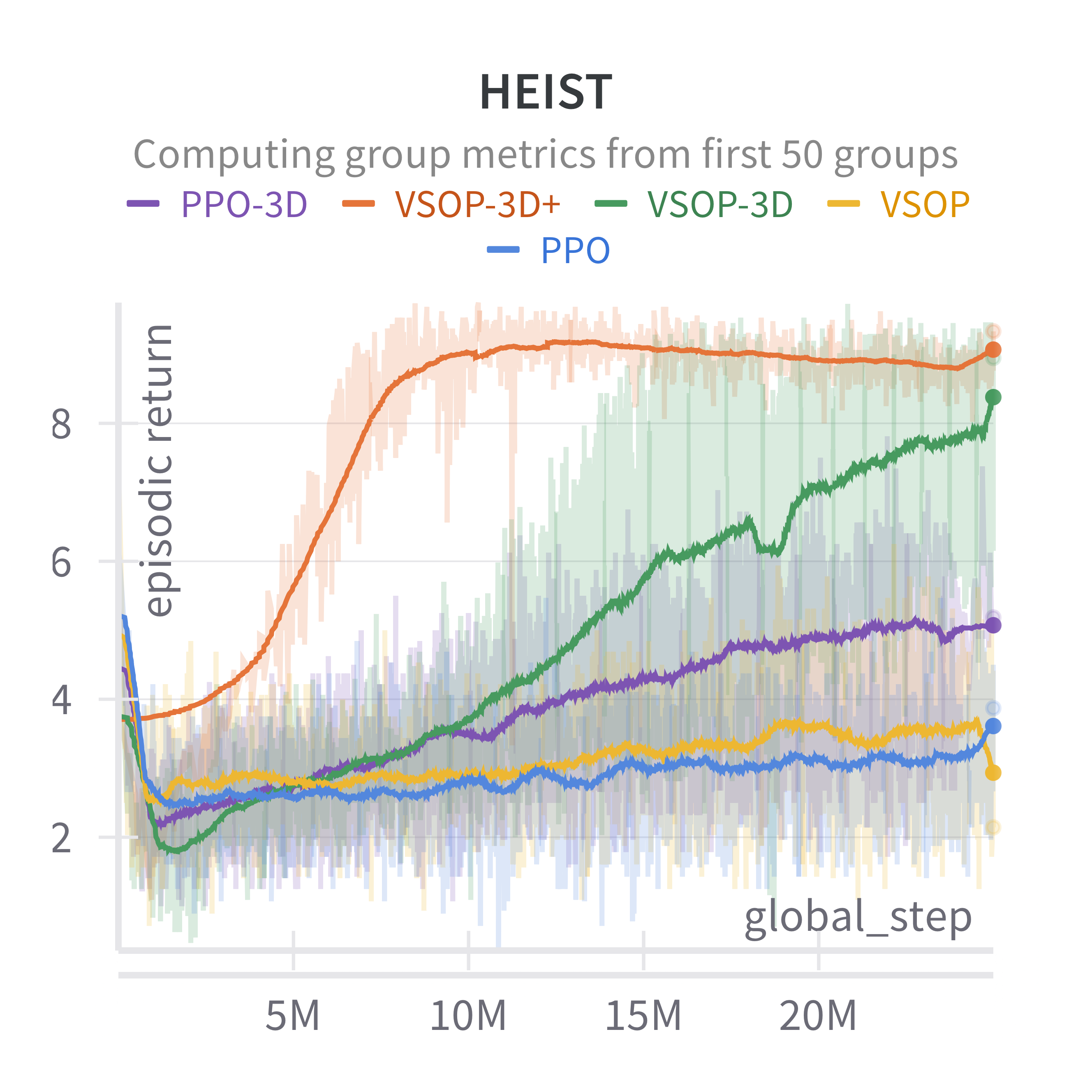} &
        \includegraphics[width=0.22\linewidth]{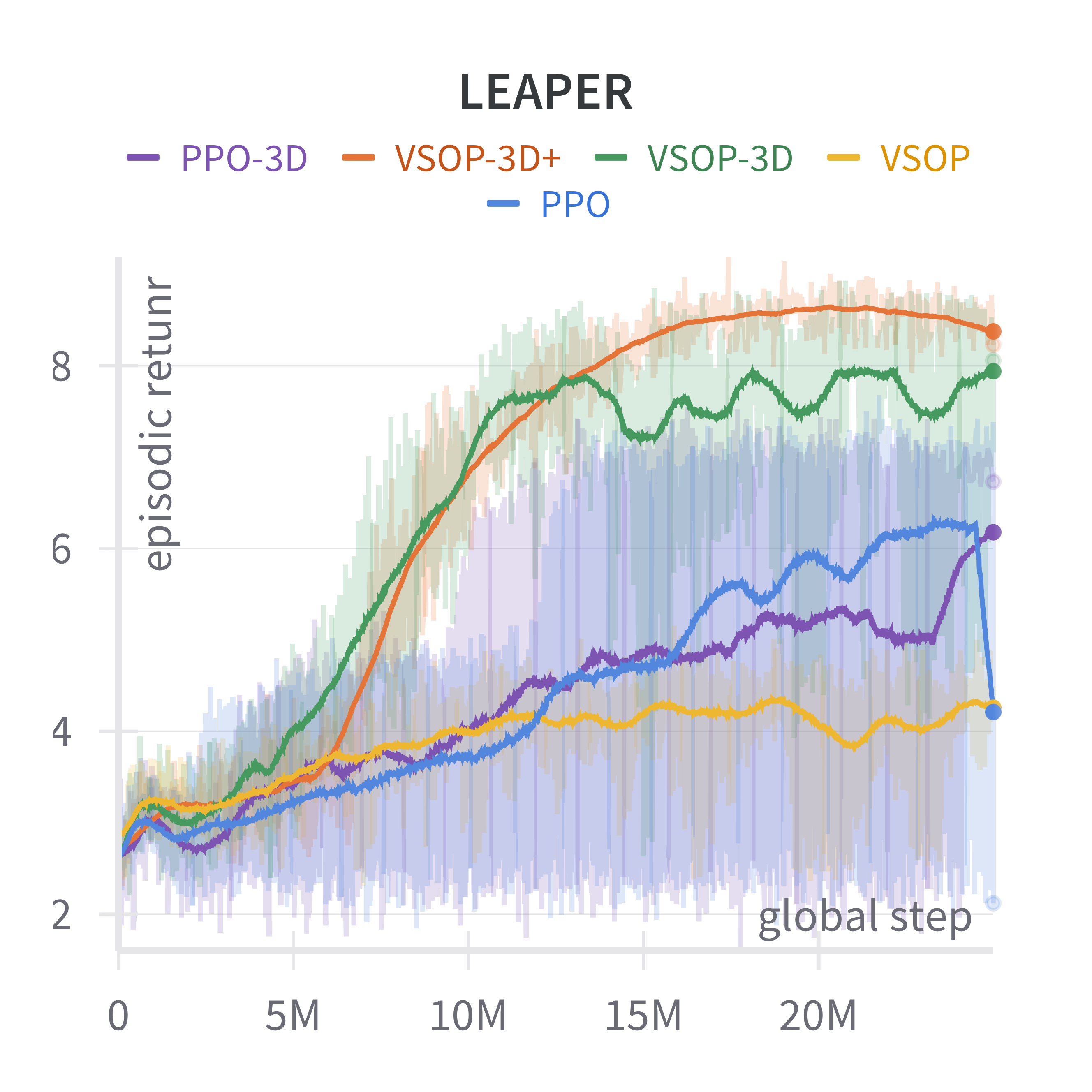} &
        \includegraphics[width=0.22\linewidth]{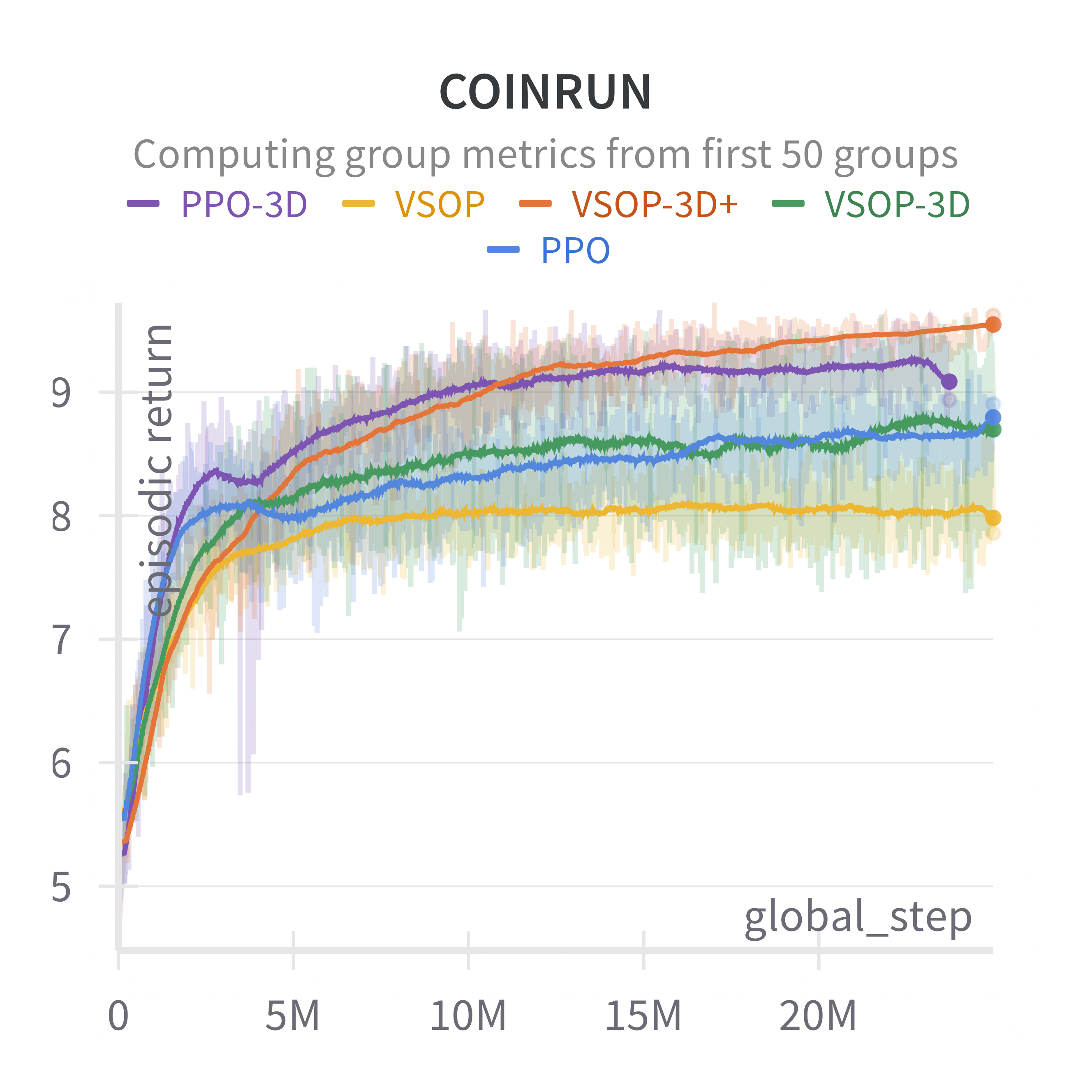} &
        \includegraphics[width=0.22\linewidth]{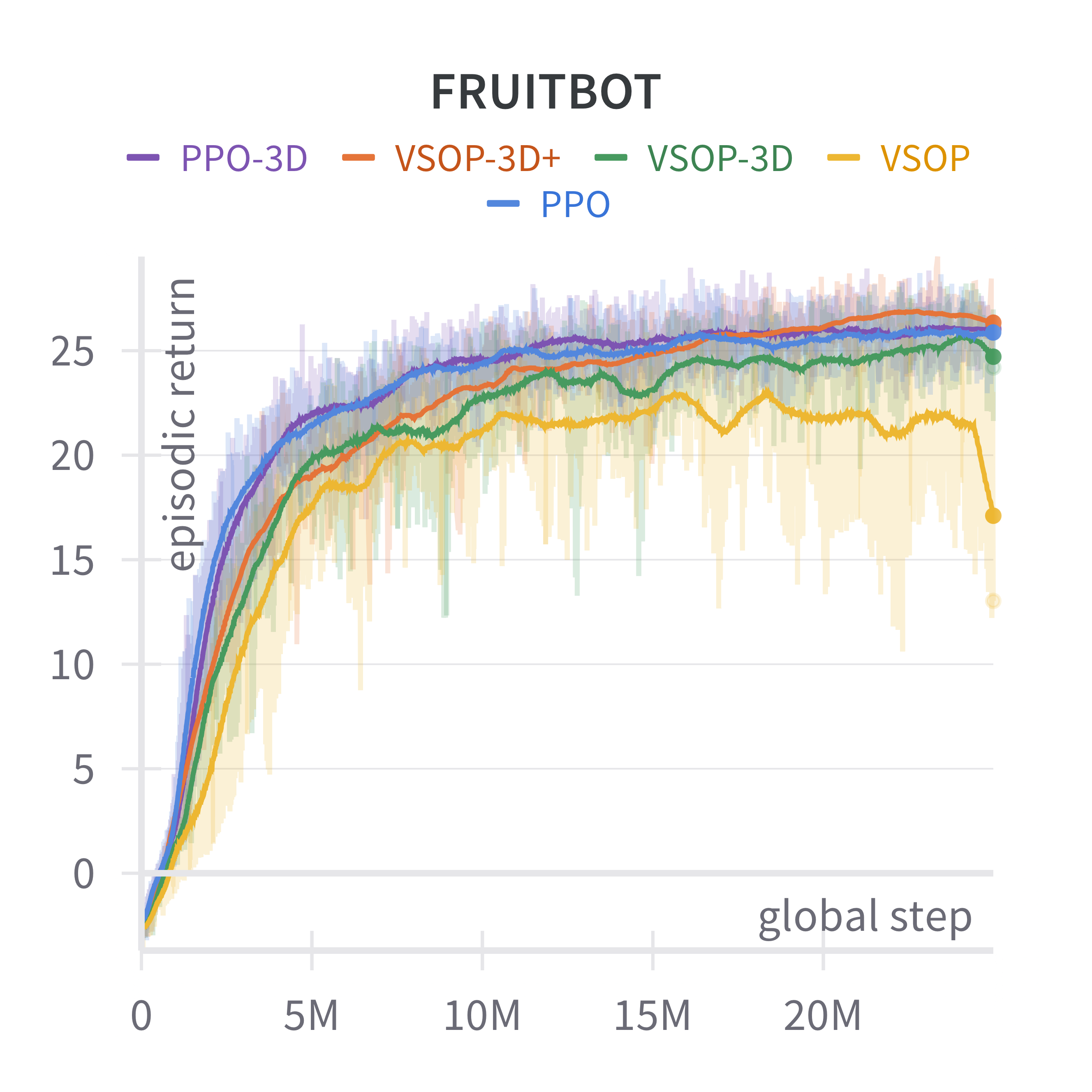} \\
        \includegraphics[width=0.22\linewidth]{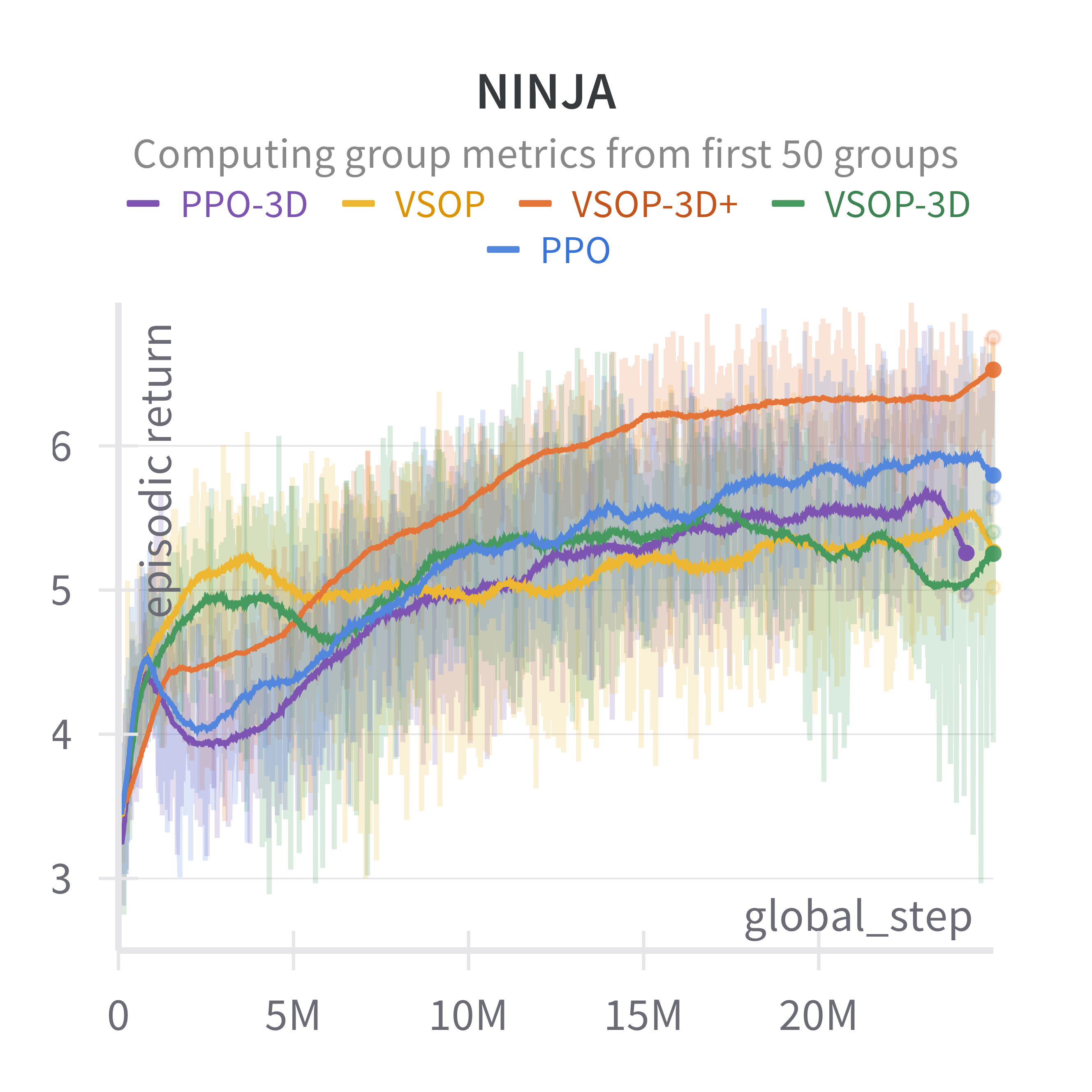} &
        \includegraphics[width=0.22\linewidth]{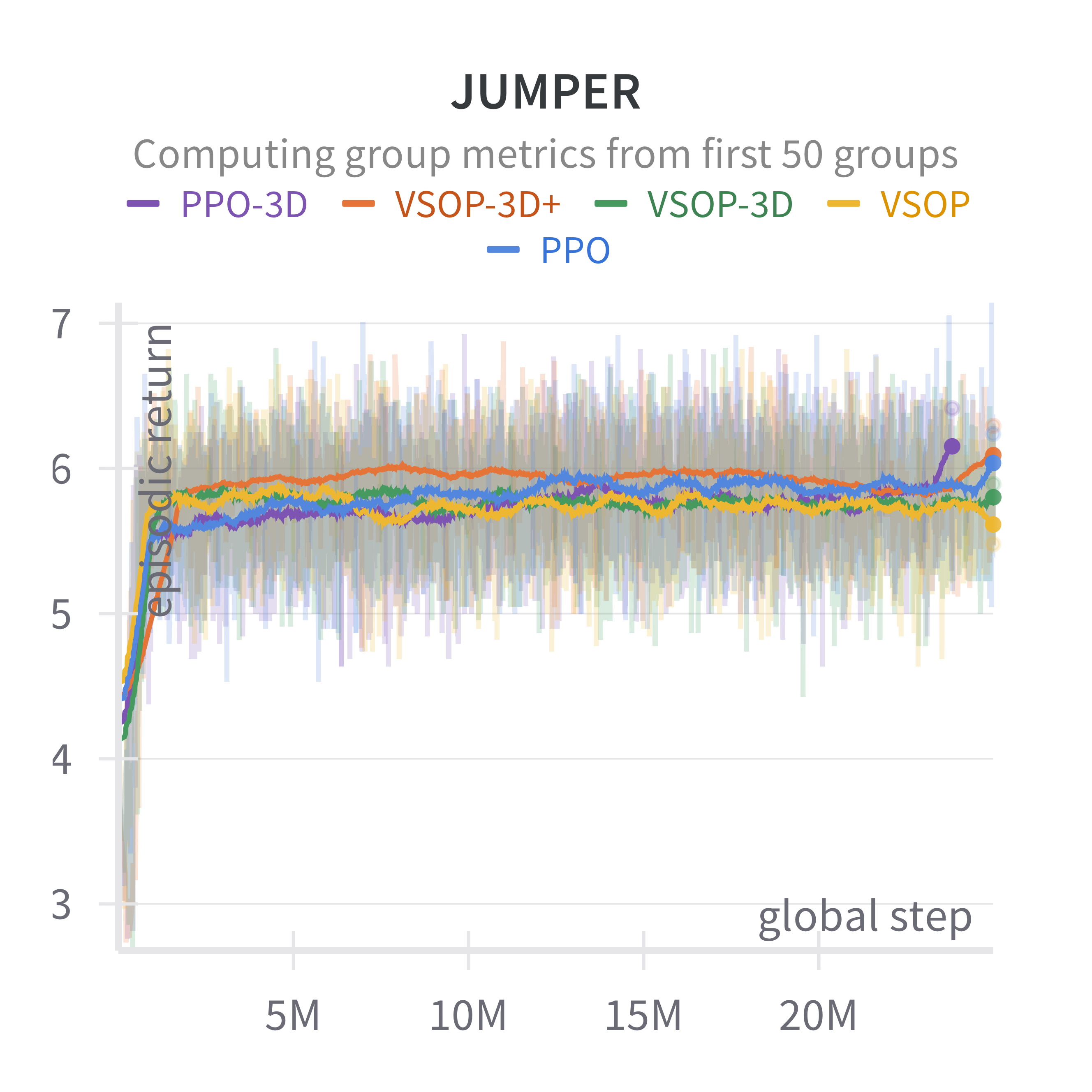} &
        \includegraphics[width=0.22\linewidth]{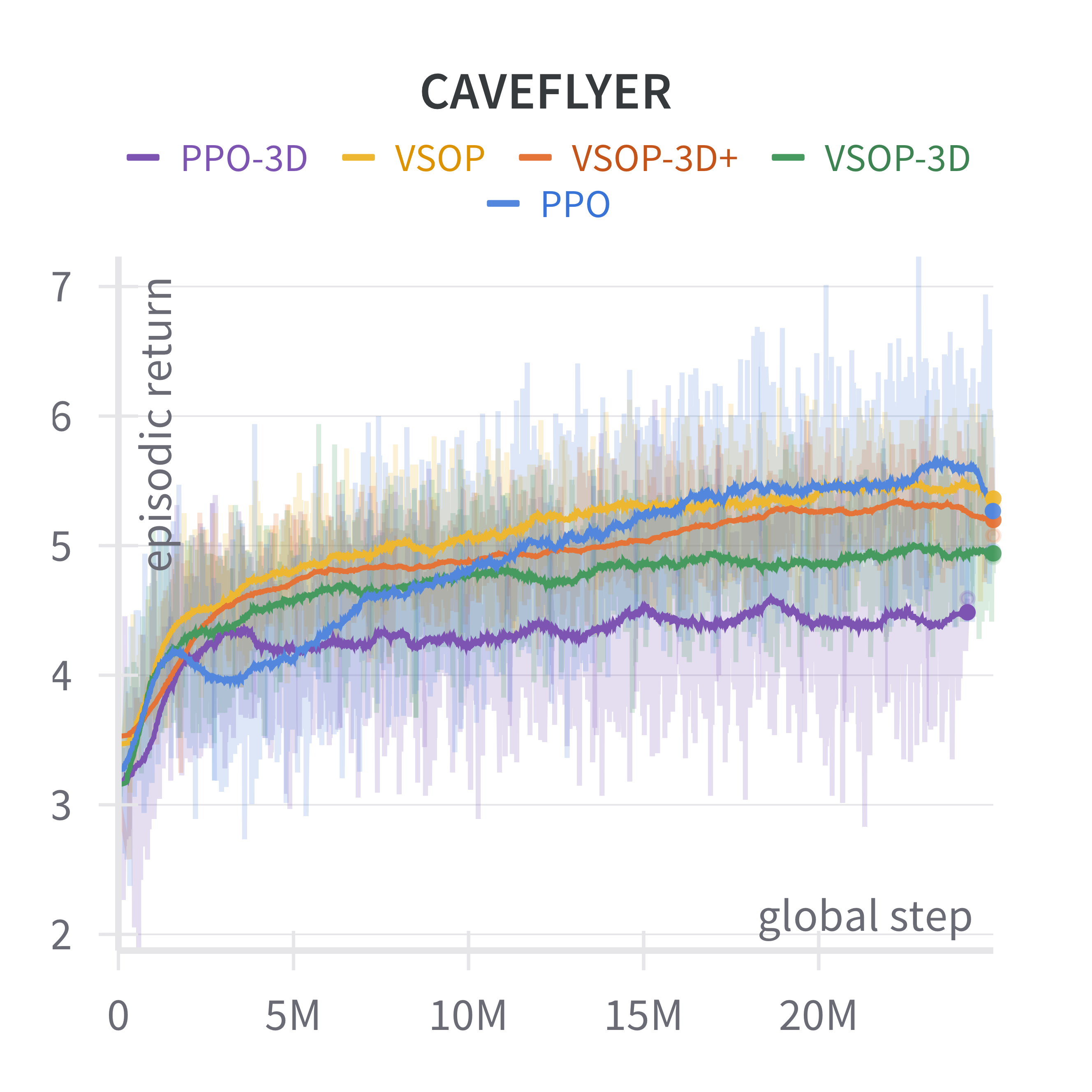} &
        \includegraphics[width=0.22\linewidth]{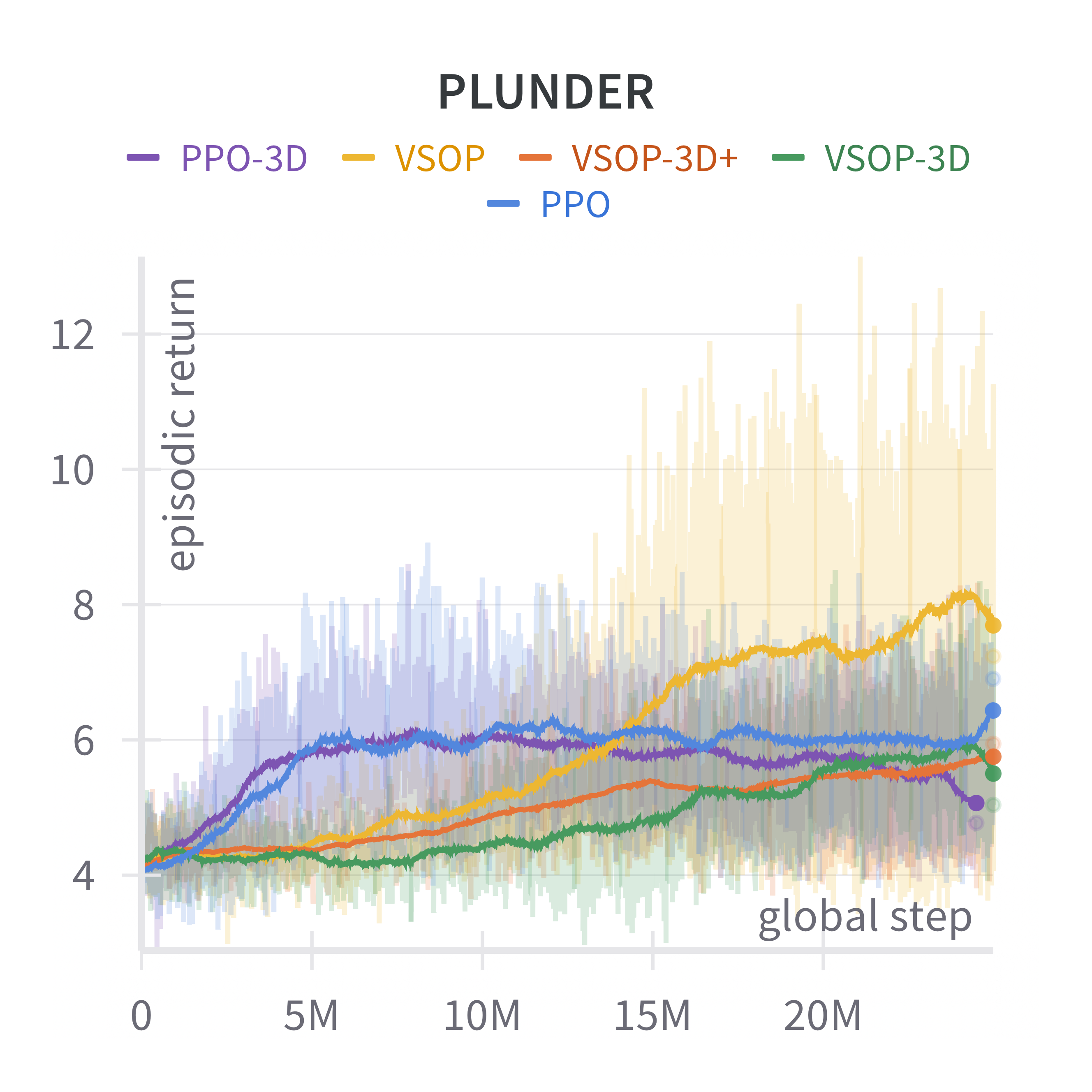} \\
    \end{tabular}
    \caption{Test episodic return curves aggregated over 5 random seeds for each ProcGen environment. VSOP-3D+ in orange, VSOP-3D in green, VSOP in yellow, PPO-3D in purple and PPO in blue. In most environments, VSOP-3D outperforms the comparisons and VSOP-3D+ improves upon VSOP-3D significantly. The only exceptions are \texttt{Jumper} and \texttt{Caveflyer} where the methods perform on par with base VSOP and \texttt{Plunger} where the proposed methods underperform base VSOP.}
    \label{fig:main}
\end{figure}

We evaluate the effect of these modifications using the ProcGen benchmark, which allows for an assessment of generalization performance by including test levels for each environment that are not used to fit the algorithm \citep{cobbe2020leveraging}.
Due to limited compute availability, we restrict our analysis to the ``easy'' difficulty setting where agents are trained on 200 levels for 25 million steps, and tested on the full distribution of levels, which is the recommended setting from \citet{cobbe2020leveraging} and the most frequently used setting in the literature.

Using the methodology of \citet{agarwal2021deep}, statistical analyses are conducted to compare the average episodic return across three different configurations of the VSOP algorithm.
The median, interquartile mean (IQM), mean, and optimality Gap metrics calculated over 5 random seeds are reported in \Cref{fig:enter-label}.
We plot the test level episodic return curves in \Cref{fig:main}.
We provide results for unmodified \textbf{VSOP} and the CleanRL \citep{huang2022cleanrl} implementation of \textbf{PPO} \citep{schulman2017proximal} as baseline references.
Due to resource constraints, we are unable to run the other baselines so we refer the readers to relevant works directly for the results~\citep{raileanu2021decoupling, zisselman2023explore}.\footnote{
    It is important to note that our performance is achieved with a \emph{single} set of hyperparameters, whereas some prior works tuned the hyperparameters for each environment.
    Our methods would likely be even better with per-environment hyperparameter tuning.
}

\textbf{Frame stacking and 3D convolutions improve generalization performance.}
Comparing VSOP-3D (green bars) and VSOP (yellow bars) in \Cref{fig:enter-label} shows that incorporating frame stacking and replacing 2D convolutions with 3D convolutions leads to a significant improvement in generalization performance, as measured by average episodic return over the last 100 steps. 
This suggests that using more temporal information (via frame stacking) and richer feature extraction (via 3D convolutions) helps improve model robustness in ProcGen environments.

\textbf{Scale improves generalization performance.}
Comparing VSOP-3D+ (orange bars) and VSOP-3D (green bars) in \Cref{fig:enter-label} shows that increasing the number of frames and the number of convolutional kernels per layer further enhances generalization, with VSOP-3D+ significantly outperforming the other configurations across all evaluation metrics.
With respect to the baseline VSOP episodic return, we observe a 65.9\% increase in the Median (from 0.44 to 0.75), a 62.8\% increase in the IQM (from 0.43 to 0.70), a 52.5\% increase in the Mean (from 0.42 to 0.64), and a 37.9\% decrease in the Optimality Gap (from 0.58 to 0.36).
This highlights the importance of scaling model capacity to better capture rich representation in the environment, which may be important for learning generalizable behavior.

\paragraph{Exploration benefits scale.} 
The benefit of scale for generalization in RL has been discussed in prior works~\citep{song2019observational}, but, intriguingly, the benefit of scale has not been widely witnessed in RL.
In most cases, naively scaling up the compute does not actually help RL.
To understand this observation better, we added the architectural change for VSOP-3D to the base PPO, which we refer to as PPO-3D.
Surprisingly, the naive application of these architectural changes did not help PPO as it did for VSOP and even hurt the performance in some environments.

We hypothesize that this observation suggests that the base PPO is \emph{data constrained}.
More concretely, PPO is not collecting useful data fast enough to benefit from increased computation scale.
In contrast, the primary objective of VSOP is to conduct better exploration.
At first glance, this statement may seem odd since both PPO and VSOP interact with the environment for the same number of steps and are trained for the same number of steps.
To resolve this paradox, it is crucial to understand how RL differs from standard supervised or unsupervised learning.
In RL, the agent needs to collect its data which means the data collected can be unhelpful if they do not provide new information.
This indicates a potential reason why scale has not helped in RL is that the main bottleneck is getting informative data that can help policy improvement rather than having enough compute to process the collected data.
In fact, scale can even hurt the performance, if the collected data are ``repeated'' -- in the sense that they only contain a small amount of new information about the environment or the optimal policy -- and the larger models are more prone to memorizing repeated data.

This hypothesis, if proven true, could have larger implications for generalization in RL beyond ProcGen or more generally interactive learning systems built on deep neural networks.
While a more thorough ablation and hyperparameter tuning are needed to understand the interaction between exploration and scaling, it is important to note that VSOP does not require any VSOP-specific hyperparameter tuning to work with increased compute scale.
In contrast, PPO does not see the same benefit.
We hope these findings will encourage the community to explore this direction further.


\section{Conclusions and future work}

Our results demonstrate that frame stacking, 3D convolutions, and scaling the number of kernels can significantly improve generalization performance in ProcGen environments. 
Specifically, our experiments show that using temporal information through frame stacking and richer spatial feature extraction through 3D convolutions results in better generalization compared to the baseline VSOP configuration. 
Furthermore, increasing the number of frames and kernels per layer (as in the VSOP-3D+ configuration) leads to additional improvements, emphasizing the importance of scaling both temporal context and model capacity.
Traditionally, deep RL has not enjoyed the benefit of scale as much compared to other areas of deep learning, so the efficacy of making the model larger is at least somewhat surprising. 
Ablation showed that the default PPO does not seem to benefit from the architecture change compared to VSOP.
We hypothesize that efficient exploration is important for leveraging scale effectively.

While these findings are promising, our analysis remains limited due to compute constraints. 
In particular, we were unable to independently investigate the effects of scaling the number of frames and scaling the number of kernels, or deeply explore how why these modifications do not yield similar performance gains when naively applied to PPO. 
We have not combined our methods with techniques such as \citet{cobbe2021phasic} and \citet{raileanu2021decoupling}. 
While we briefly investigated using transformer-based architectures~\citep{vaswani2017attention,parisotto2020stabilizing} instead of 3D convolutions, the initial results were not compelling suggesting that further research or hyperparameter tuning is likely required.
Finally, there may be a connection between incorporating temporal convolution and using structured state space sequence models \citep{gu2022efficiently,smith2023simplified,lu2024structured} for reinforcement learning that could be explored.
Conducting a thorough analysis will require significant computational resources, which were not available for this study.

To enable further research, we have made our code publicly available at \url{https://github.com/anndvision/vsop-3d}.
We hope this will facilitate replication of our results and allow future work to understand the effects of scale and exploration in more depth, providing a more complete picture of how these modifications impact both generalization and optimization in reinforcement learning tasks.

\bibliographystyle{unsrtnat}
\bibliography{main}

\end{document}